\documentclass[preprint,12pt,authoryear]{elsarticle}

\usepackage{booktabs}

\usepackage{amsmath}

\usepackage{algorithm}
\usepackage{algorithmic}
\usepackage{subfig}

\usepackage{amsfonts}
\usepackage{amssymb}
\usepackage{amsmath}
\usepackage[export]{adjustbox}
\usepackage{graphicx}
\usepackage{multirow}
\usepackage{tabularx}
\usepackage{pifont} 
\usepackage{makecell}
\usepackage[colorlinks]{hyperref}

\usepackage{cleveref}
\journal{Journal of Visual Communication and Image Representation}

\begin{document}

\begin{frontmatter}
\title{Hierarchical mask-enhanced dual reconstruction network for few-shot fine-grained image classification}

\author[label1]{Ning Luo%
	\fnref{fn1}}

\author[label1]{Meiyin Hu%
	\fnref{fn1}}

\author[label2]{Huan Wan}

\author[label3]{Yanyan Yang}

\author[label4]{Zhuohang Jiang}

\author[label1]{Xin Wei%
	\corref{cor1}}%
\ead{xinwei@ncu.edu.cn}

\affiliation[label1]{
	organization={School of Software, Nanchang University},
	city={Nanchang},
	postcode={330047},
	state={Jiangxi},
	country={China}
}

\affiliation[label2]{
	organization={School of Artificial Intelligence, Jiangxi Normal University},
	city={Nanchang},
	postcode={330022},
	state={Jiangxi},
	country={China}
}

\affiliation[label3]{
	organization={School of Software Engineering, Beijing Jiaotong University},
	city={Beijing},
	postcode={100044},
	country={China}
}

\affiliation[label4]{
	organization={School of Electronic Information and Communications, Huazhong University of Science and Technology},
	city={Wuhan},
	postcode={430070},
	state={Hubei},
	country={China}
}

\cortext[cor1]{Corresponding author.}
\fntext[fn1]{These authors contributed equally to this work.}

\begin{abstract}
Few-shot fine-grained image classification (FS-FGIC) is challenging as it requires distinguishing visually similar subclasses with extremely limited labeled examples. Existing methods suffer from critical limitations: metric-based methods lose spatial information and misalign local features, while reconstruction-based methods underuse hierarchical feature information and lack selective focus on discriminative key regions. We propose the Hierarchical Mask-enhanced Dual Reconstruction Network (HMDRN), integrating dual-layer feature reconstruction with mask-enhanced feature processing. HMDRN leverages complementary visual information from different network hierarchies via learnable weights, balancing high-level semantic representations with mid-level structural details. It incorporates a spatial binary mask-enhanced transformer module that selectively enhances discriminative regions while filtering background noise. On three fine-grained datasets, HMDRN consistently outperforms state-of-the-art methods with both Conv-4 and ResNet-12 backbones. Ablation studies validate each component's effectiveness, showing dual-layer reconstruction enhances inter-class discrimination while mask-enhanced transformation reduces intra-class variations. Code is available at:  ~\href{https://github.com/2004-0813/HMDRN-main}{\textcolor{blue}{https://github.com/2004-0813/HMDRN-main}}.
\end{abstract}


\end{frontmatter}

\section{Introduction}
Few-shot learning has emerged as a critical research direction in computer vision, aiming to equip models with the ability to generalize from very few labeled examples \citep{ZHENG2019563}. Within this paradigm, the task of few-shot fine-grained image classification (FS-FGIC) introduces additional complexity, as it requires distinguishing between visually similar subcategories under data-scarce conditions \citep{CHEN2022103678}. Unlike general few-shot learning, FS-FGIC demands the capability to capture subtle visual differences that differentiate closely related subclasses, such as distinct bird species with similar plumage. This task is particularly challenging because discriminative features are often confined to specific regions and exhibit high intra-class variability alongside low inter-class differences, a difficulty also observed in other fine-grained recognition scenarios \citep{ZHANG2024104172,ding2024multidimensional}. Consequently, models must learn to focus on discriminative details yet maintain robust generalization capabilities with limited labeled examples.

Recent advances in addressing FS-FGIC primarily follow two types of methods: metric-based methods and reconstruction-based methods. Metric-based methods learn embeddings that minimize intra-class distances while maximizing inter-class separation. ProtoNet \citep{snell2017prototypical} computes class prototypes and classifies samples based on their Euclidean distances to the prototypes. MatchingNet \citep{DBLP:conf/nips/VinyalsBLKW16} employs attention mechanisms to generate weighted combinations of support samples for classification. RelationNet \citep{sung2018learning} utilizes neural networks to compute similarity scores between query and support features. Other methods focus on local information: DN4 \citep{8953758} introduces an image-to-class measure based on local descriptors, BSNet \citep{li2020bsnet} implements multiple similarity metrics to align discriminative local features, and LR-PABN \citep{9115215} employs bilinear pooling to align discriminative regions. Despite these efforts to capture local details, a fundamental limitation of these methods is that they transform three-dimensional feature maps into flattened vectors, which inevitably results in spatial information loss and misalignment of the discriminative local features that are essential for distinguishing between visually similar subclasses.

Reconstruction-based methods have emerged as promising alternatives that partially address the spatial information limitations of metric-based methods by preserving feature map structures during the forward propagation of the network. FRN \citep{wertheimer2021few} uses ridge regression to reconstruct query features from support features and determines the class to which it belongs by measuring reconstruction error. BiFRN~\citep{li2023bidirectional} extends FRN through bidirectional reconstruction to simultaneously enhance inter-class discrimination and intra-class variations. LCCRN \citep{li2023locally} develops a specialized local content extraction module designed to capture discriminative local information. Recent innovations include channel-wise reconstruction \citep{yang2024channel}, query-focused methods \citep{zhang2024query}, and self-reconstruction techniques \citep{li2024self} that enhance feature discrimination. However, despite their advantages, the reconstruction-based methods still exhibit two critical limitations: first, they only use the output of the last layer of the network as the extracted features, ignoring the complementarity of visual information embedded across different network hierarchies; second, they lack effective mechanisms to dynamically isolate the most discriminative regions from background noise and redundant information, which is particularly problematic when distinguishing between highly similar subclasses that differ only in subtle details.These limitations significantly constrain the discriminative power of feature representations in FS-FGIC tasks.

\begin{figure}[ht]
	\centering
	\includegraphics[width=0.9\linewidth]{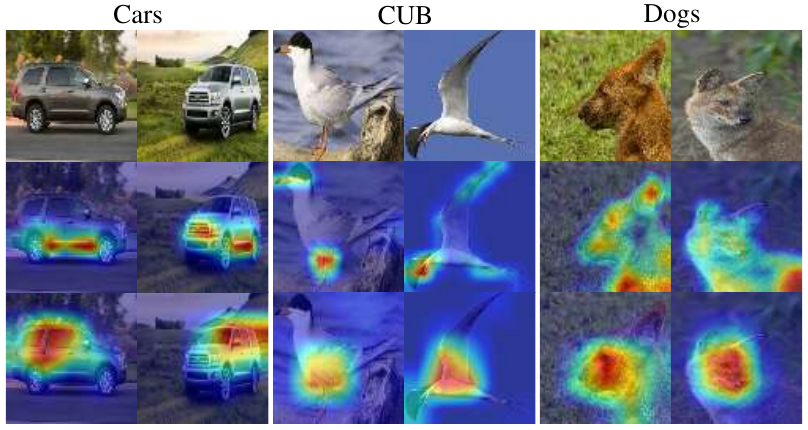}
	\caption{Feature visualization extracted by backbone networks across Stanford-Cars \citep{krause20133d}, CUB-200-2011 \citep{WelinderEtal2010}, and Stanford-Dogs \citep{xiao2011sun} datasets. The first row presents original input images, the second row displays feature activations from the penultimate layer of the backbone, and the third row illustrates activations from the last layer. Regions with higher color intensity indicate stronger feature activations, highlighting areas where the network responds more prominently during feature extraction.}
	\label{figure_1}
\end{figure}

To address these limitations, our analysis reveals two key insights. First, our systematic investigation of reconstruction-based feature hierarchies reveals a previously unexplored spatial-semantic complementarity between the penultimate and last layers of deep neural networks for fine-grained tasks. As shown in Fig. \ref{figure_1}, the penultimate layer exhibits more dispersed activation distributions, capable of capturing broader regional features and local discriminative features of objects. In contrast, the last layer displays more focused activation patterns, primarily concentrated on the most discriminative core regions, providing higher-level semantic representations. This complementarity forms an ideal feature combination for fine-grained recognition, enabling comprehensive analysis of targets at different abstraction levels. 

Second, we find that traditional methods based on continuous attention weights face significant limitations in fine-grained scenarios where enhancing focus on discriminative regions is crucial. As shown in Fig. \ref{figure_1}, feature extraction processes retain low but non-zero weights in background regions, which increases classification uncertainty when inter-class differences are minimal. These continuous distributions dilute discriminative information by incorporating less informative features. Our proposed spatial binary mask strategy constructs a hard-decision mechanism that divides the feature space into two discrete states through adaptive thresholding, eliminating less informative regions. This discretization improves fine-grained classification by creating higher-contrast representations, reducing feature space complexity to mitigate overfitting, and enhancing the clarity of decision boundaries between visually similar subclasses.

Based on the above two insights, we propose the Hierarchical Mask-enhanced Dual Reconstruction Network (HMDRN) for few-shot fine-grained image classification. The HMDRN extracts dual-layer feature maps from the backbone network, then processes them through two innovative modules: (1) Cross-Level Attentional Reconstruction Module (CLARM), which integrates complementary information from different feature hierarchies to enhance representational capacity; (2) Masked Transformer Feature Enhancement Module (MTFEM), which applies adaptive binary masks to selectively process discriminative regions. The complete architecture sequentially enhances features through MTFEM, reconstructs each layer separately, and adaptively combines similarity scores using learnable weights.

In summary, the main contributions of our work are as follows:
\begin{itemize}
	\item We propose the Hierarchical Mask-enhanced Dual Reconstruction Network (HMDRN) for few-shot fine-grained image classification, which addresses the dual challenges of utilizing complementary information across network hierarchies and enhancing focus on discriminative regions;
	\item We design a Cross-Level Attentional Reconstruction Module (CLARM) that addresses the underutilization of hierarchical features by effectively utilizing and integrating complementary visual representations from different network layers, simultaneously capturing high-level semantic representations and mid-level structural details, thus resolving the feature expression bottleneck in FS-FGIC tasks;
	\item We develop a novel Masked Transformer Feature Enhancement Module (MTFEM) with an adaptive binary masking strategy that addresses the lack of selective region focus by enhancing focus on discriminative regions while filtering less informative features, improving the ability of the model to distinguish subtle differences between similar subclasses; Operating sequentially, MTFEM refines features for CLARM to reconstruct across hierarchical levels;
	\item We conduct extensive experiments on three challenging fine-grained datasets, with ablation studies and feature visualizations demonstrating our method consistently outperforms state-of-the-art methods while providing visual evidence of superior feature reconstruction capabilities.
\end{itemize}

\section{Related work}
\label{Related Work}
\subsection{Metric-based few-shot learning}

Metric-based methods constitute a fundamental paradigm in few-shot learning by learning embedding spaces where similarity relationships between classes can be effectively quantified. BSNet \citep{li2020bsnet} implements a dual-similarity architecture that optimizes intra-class compactness and inter-class separability, enhancing discriminative local feature alignment. LR-PABN \citep{9115215} advanced the field by employing bilinear pooling to align subtle discriminative regions across images. TempNet \citep{ZHU2021107797} introduced an adaptive method using Gaussian kernel functions for query-specific spatial optimization, enabling tailored similarity measurements. NDPNet \citep{Zhang2021NDPNetAN} enhanced discrimination capability through nonlinear projection mechanisms, incorporating the sign function to create sharper decision boundaries between visually similar classes. More recently, PHR \citep{9767592} developed a multi-level representation framework that integrates spatial, global, and semantic features, improving prototype calibration and inter-class discriminability. Despite these significant advances, metric-based methods inherently face challenges with spatial information preservation when transforming feature maps into vector representations, thus limiting their efficacy for tasks requiring precise spatial alignment of discriminative features.

\subsection{Reconstruction-based few-shot learning}

Reconstruction-based methods address the spatial information preservation challenge by maintaining structural integrity throughout the feature processing pipeline. FRN \citep{wertheimer2021few} established the method of using ridge regression to reconstruct query features from support features, employing reconstruction error as a class membership criterion. BiFRN \citep{li2023bidirectional} introduced a bidirectional mechanism that simultaneously reconstructs query features from support features and vice versa, addressing both inter-class and intra-class variations. LCCRN \citep{li2023locally} further advanced this direction by developing specialized modules for extracting discriminative local content within feature maps. Recent innovations have expanded this paradigm through complementary strategies: self-reconstruction techniques \citep{li2024self}, query-focused methods \citep{zhang2024query}, and center-based methods \citep{li2024simple}. While these methods successfully preserve spatial information, they predominantly extract features from only a single network layer, overlooking the rich complementary information embedded across different hierarchical levels, which is the limitation our method specifically addresses.

\subsection{Attention mechanisms in few-shot learning}
Attention mechanisms have revolutionized few-shot learning by enabling selective focus on discriminative features crucial for fine-grained recognition. Seminal contributions include FEAT \citep{ye2020few}, which employs Transformer-based set-to-set functions for dynamic embedding adaptation, and CTX \citep{doersch2020crosstransformers}, which identifies spatial correspondences between images through self-attention. The AIM method \citep{9711099} implemented attention-driven expert mixture models for selective feature activation. \citep{lee2022task} specialized attention for fine-grained tasks with TDM, introducing support and query attention modules to locate discriminative regions through task-specific channel weighting. Recent advancements include \citep{yang2024channel}, who proposed CSCAM, integrating channel and spatial dimensions to capture cross-information between support and query samples, while the research presented in \citep{10763467} extended TDM with an Instance Attention Module for intermediate feature layers. Although these methods demonstrate significant progress, the integration of adaptive masking with Transformer-based attention across multiple network layers remains underexplored. HMDRN advances beyond these methods by unifying dual-layer feature reconstruction with binary mask-enhanced transformation, addressing spatial information loss, single-layer limitation, and soft-attention redundancy in a single framework for improved fine-grained discrimination. Our method filled this research gap by combining a mask-enhanced Transformer module with a multi-layer feature reconstruction strategy.

\begin{figure*}[!ht]
	\centering
	\includegraphics[width=1.0\linewidth]{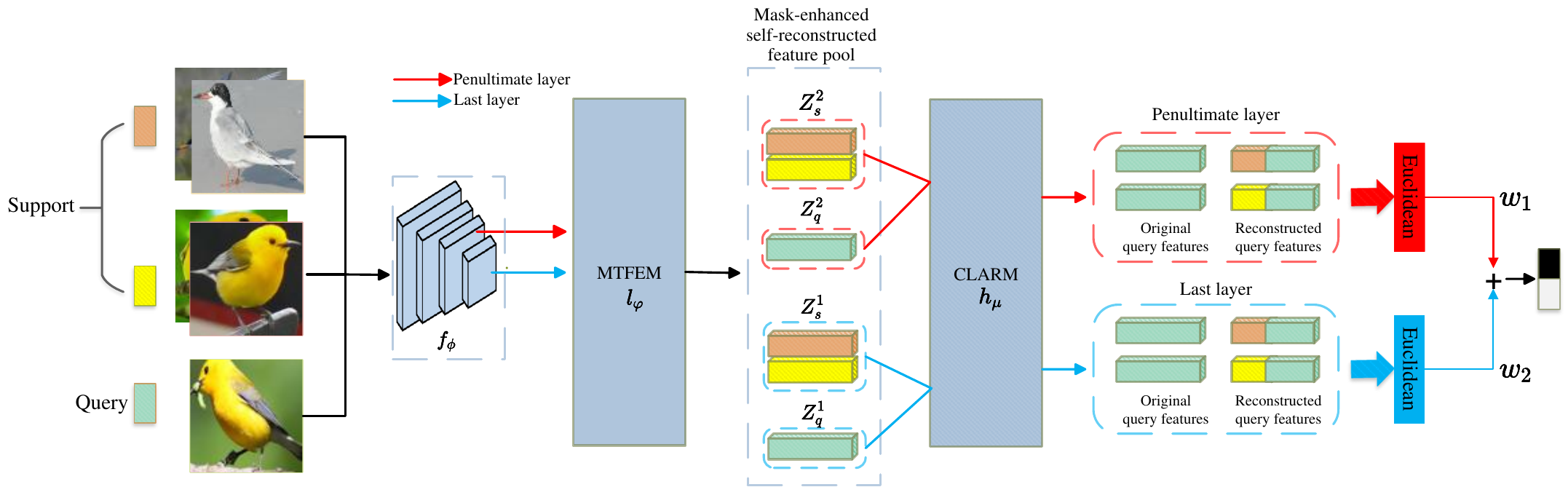}
	\caption{The architecture of the HMDRN. Orange and yellow blocks represent two subclasses of support set images, while green blocks represent query set images. Input images are first processed through a backbone network to extract features from the penultimate layer (red pathways) and the last layer (blue pathways). These dual-level features are then processed by a mask-enhanced self-reconstruction module that focuses attention on discriminative regions. Subsequently, both feature levels are processed through separate reconstruction networks, generating two sets of reconstruction results. Finally, the similarity scores from both hierarchical levels are combined through learnable weights ($w_1$ and $w_2$) to produce the final classification score.}
	\label{figure_2}
\end{figure*}

\section{Methodology}
\label{Methodology}
In this section, we describe our proposed method for few-shot fine-grained image classification. We start with the problem formulation, followed by an overview of the architecture and a detailed description of each component.

\subsection{Problem Definition}
\label{Problem Definition}
Given a dataset $D = \{(x_i, y_i), y_i \in Y\}$, where $x_i$ represents the original image and $y_i$ denotes its corresponding class label, we partition it into three mutually exclusive subsets: training set $D_{train} = \{(x_i, y_i), y_i \in Y_{train}\}$, validation set $D_{val} = \{(x_i, y_i), y_i \in Y_{val}\}$, and test set $D_{test} = \{(x_i, y_i), y_i \in Y_{test}\}$. These subsets are mutually disjoint, satisfying $Y_{train} \cap Y_{val} \cap Y_{test} = \emptyset$ and $Y_{train} \cup Y_{val} \cup Y_{test} = Y$.

The meta-learning framework and episodic training strategy are adopted. The meta-learning framework establishes a principled correspondence between the learning phase on training set $D_{train}$ and subsequent inference on test set $D_{test}$, thereby facilitating effective knowledge transfer to novel fine-grained classes despite limited samples. The episodic training strategy enhances model adaptability by systematically simulating realistic few-shot classification scenarios throughout the learning process. During the training phase, each episode comprises $N$ classes randomly selected from training set $D_{train}$, then extracting $K$ exemplar images per class to constitute the support set, alongside $M$ additional images from the same classes to form the query set, thereby instantiating $N$-way $K$-shot classification tasks.

The fundamental objective of FS-FGIC is to accurately classify query instances in $N$-way $K$-shot tasks formulated from test set $D_{test}$ by leveraging the knowledge acquired from training set $D_{train}$. The model undergoes optimization across multiple $N$-way $K$-shot episodes sampled from the training set $D_{train}$, with parameter refinement guided by classification loss minimization. The validation set $D_{val}$ serves as the criterion for determining optimal model configuration, while comprehensive performance assessment occurs on tasks derived from the test set $D_{test}$ to quantify generalization capabilities across previously unseen fine-grained classes.

\subsection{Overview of the proposed method}
\label{Overview of the Proposed Method}
Fig. \ref{figure_2} illustrates the overall architecture of our proposed Hierarchical Mask-Enhanced Dual Reconstruction Network (HMDRN).
The embedding module $f_{\phi}$ extracts dual-level feature representations from both the penultimate and last layers of the backbone network, capturing visual information at different abstraction levels.
The Masked Transformer Feature Enhancement Module $l_{\varphi}$ employs a self-attention mechanism where the convolutional features of each image are reconstructed by themselves, with a spatial binary mask strategy applied specifically to the query set.
This adaptive masking mechanism highlights the most discriminative key regions in images while effectively filtering background noise and redundant information.
The Cross-Level Attentional Reconstruction Module $h_{\mu}$ reconstructs query features based on support set features, executing this process in parallel across two different levels of feature representation, thereby leveraging complementary information from different hierarchies.
Finally, the Euclidean metric module computes the distance between original query features and their reconstructed counterparts, and adaptively combines the similarity scores from both feature levels through learnable parameters ($w_1$ and $w_2$) to generate the final classification result.
This method balances comprehensive feature representation with precise focus on discriminative regions, addressing the fundamental challenges of differentiating visually similar subclasses with limited examples.

\subsection{Feature Extraction}
\label{Feature Extraction}

Our feature extraction module implements a dual-level representation strategy by extracting features from two distinct network hierarchies. Unlike conventional methods that rely solely on the output from a single layer, we capture complementary information from different network depths within the backbone architecture. Formally, given an input image $x$, our feature extraction process can be represented as:
\begin{equation}
	f_{high}, f_{mid} = f_{\phi}(x),
\end{equation}
\noindent where $f_{\phi}$ denotes the backbone network, and $f_{high} \in \mathbb{R}^{C_1 \times r_1 \times r_1}$ and $f_{mid} \in \mathbb{R}^{C_2 \times r_2 \times r_2}$ represent the extracted feature maps from the last and penultimate layers, respectively. The channel dimensions $C_1$ and $C_2$ and spatial resolutions $r_1$ and $r_2$ vary according to the specific backbone architecture.

This dual-level extraction strategy transcends the common single-layer limitation prevalent in existing methods, providing richer and more diverse visual information for the subsequent reconstruction processes.

\subsection{Masked Transformer Feature Enhancement Module}
\label{Masked Transformer Feature Enhancement Module (MTFEM)}
\begin{figure}[ht]
	\centering
	\includegraphics[width=0.8\linewidth]{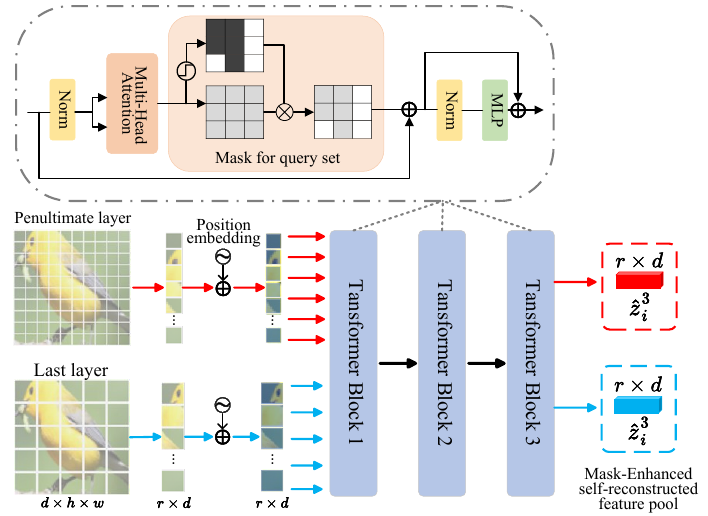}
	\caption{Masked Transformer Feature Enhancement Module.}
	\label{figure_3}
\end{figure}
In our HMDRN, the Masked Transformer Feature Enhancement Module (MTFEM) implements an adaptive binary masking strategy that selectively filters features based on self-attention scores, enhancing focus on discriminative regions crucial for fine-grained classification, as illustrated in Fig. \ref{figure_3}.

Given the dual-level feature maps extracted by the embedding module for a $N$-way $K$-shot classification task, we reshape each feature map $\hat{x}_i \in \mathbb{R}^{r \times d}$ into a sequence of local features at spatial positions $[\hat{x}_i^1, \hat{x}_i^2, ..., \hat{x}_i^r]$, where $r = h \times w$ represents the spatial dimension ($r_1$ for last-layer features and $r_2$ for penultimate-layer features) and $d$ represents the feature channel dimension. To create position-aware feature representations, we add positional embedding $E_{pos} \in \mathbb{R}^{r \times d}$ to the sequence. This results in our position-encoded feature sequence $z_i$:
\begin{equation}
	z_i = [\hat{x}_i^1, \hat{x}_i^2, ..., \hat{x}_i^r] + E_{pos},
\end{equation}
where $E_{pos}$ employs sinusoidal position encoding to preserve spatial relationships.

The core innovation of our MTFEM lies in the Masked Attention operation with binary masking. The multi-head self-attention mechanism with two heads is defined as:
\begin{equation}
	\operatorname{MSA}(z_i) = \operatorname{Concat}(\text{head}_1, \text{head}_2) W^O,
	\label{equation_msa}
\end{equation}
where $\text{head}_1$ and $\text{head}_2$ operate independently on different representation subspaces, and $W^O \in \mathbb{R}^{2d \times d}$ projects the concatenated results back to the original dimension. Both heads independently perform the attention operations described below.

Standard self-attention operation is defined as:
\begin{equation}
	\operatorname{Attention}(Q, K, V) = \operatorname{Softmax}\left(\frac{Q K^T}{\sqrt{d_k}}\right)V.
	\label{equation_attention}
\end{equation}
For an input feature $z_i$, we project it into query, key, and value representations:
\begin{equation}
	q = z_iW^Q, \quad k = z_iW^K, \quad v = z_iW^V,
\end{equation}
where $W^Q$, $W^K$, and $W^V \in \mathbb{R}^{d \times d}$ are learnable weight matrices. For query features specifically, we implement our spatial binary masking mechanism to compute masked query representations $q'$ as follows:
\begin{equation}
	Q_{scores} = \|q\|_2^2 \in \mathbb{R}^r, \quad
	\bar{Q}_{score} = \frac{1}{r}\sum_{i=1}^{r} (Q_{scores})_i,
\end{equation}

\begin{equation}
	M_q = \delta(Q_{scores} > \bar{Q}_{score}),
\end{equation}

\begin{equation}
	q' = q \odot M_q,
\end{equation}
\noindent where $\odot$ denotes element-wise multiplication. $\delta(\cdot)$ is the indicator function that outputs 1 when the condition is satisfied, and 0 otherwise. This mechanism generates a hard-threshold binary mask that retains only the most informative query representations. Each attention head computes its mask independently, allowing focus on different informative regions.

For query features, the self-attention process with masking produces output $\hat{z}_i$ as:
\begin{equation}
	\hat{z}_i = \text{Attention}(q',k,v) = \text{Softmax}\left(\frac{q'k^T}{\sqrt{d_k}}\right)v.
\end{equation}
For support features, we maintain standard self-attention without masking to generate output $\hat{z}_i$ as:
\begin{equation}
	\hat{z}_i = \text{Attention}(q,k,v) = \text{Softmax}\left(\frac{qk^T}{\sqrt{d_k}}\right)v.
\end{equation}
Our MTFEM employs a three-layer Transformer encoder with two attention heads per layer. For the first encoder layer, given input sequence $z_i$, the output $\hat{z}_i^1$ is computed as:
\begin{equation}
	\begin{aligned}
		\hat{z}_i^1 &= \operatorname{MLP}\left(\operatorname{LN}\left(\operatorname{MSA}\left(\operatorname{LN}(z_i)\right) + z_i\right)\right) \\
		&\quad + \operatorname{MSA}\left(\operatorname{LN}(z_i)\right) + z_i, \quad \hat{z}_i^1 \in \mathbb{R}^{r \times d},
	\end{aligned}
\end{equation}
where $\operatorname{MSA}$ denotes Multi-Head Self-Attention (with masking for query features), $\operatorname{LN}$ represents Layer Normalization, and $\operatorname{MLP}$ is a Multi-Layer Perceptron with GELU activation.

For the second encoder layer, with input $\hat{z}_i^1$ obtained from the first layer, the output $\hat{z}_i^2$ is computed as:
\begin{equation}
	\begin{aligned}
		\hat{z}_i^2 &= \operatorname{MLP}\left(\operatorname{LN}\left(\operatorname{MSA}\left(\operatorname{LN}(\hat{z}_i^1)\right) + \hat{z}_i^1\right)\right) \\
		&\quad + \operatorname{MSA}\left(\operatorname{LN}(\hat{z}_i^1)\right) + \hat{z}_i^1, \quad \hat{z}_i^2 \in \mathbb{R}^{r \times d}.
	\end{aligned}
\end{equation}
Similarly, the third encoder layer takes $\hat{z}_i^2$ from the second layer as input, and computes the output $\hat{z}_i^3$ as:
\begin{equation}
	\begin{aligned}
		\hat{z}_i^3 &= \operatorname{MLP}\left(\operatorname{LN}\left(\operatorname{MSA}\left(\operatorname{LN}(\hat{z}_i^2)\right) + \hat{z}_i^2\right)\right) \\
		&\quad + \operatorname{MSA}\left(\operatorname{LN}(\hat{z}_i^2)\right) + \hat{z}_i^2, \quad \hat{z}_i^3 \in \mathbb{R}^{r \times d}.
	\end{aligned}
\end{equation}

This binary thresholding method enhances fine-grained classification by creating discrete feature representations that completely eliminate less informative regions. This discretization improves inter-class separation while reducing feature space complexity, creating clearer decision boundaries for visually similar subclasses.

\subsection{Cross-Level Attentional Reconstruction Module}
\label{sec:CLARM}

\begin{figure}[ht]
	\centering
	\includegraphics[width=0.9\linewidth]{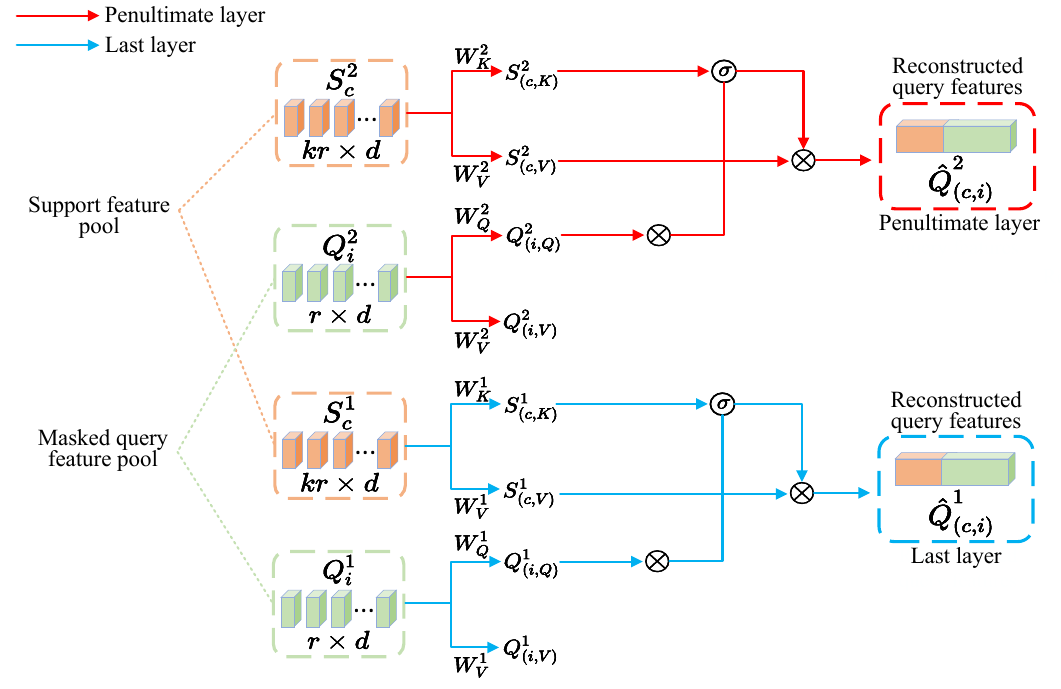}
	\caption{Cross-Level Attentional Reconstruction Module performing feature reconstruction at dual hierarchical levels.}
	\label{figure_4}
\end{figure}

The Cross-Level Attentional Reconstruction Module (CLARM), illustrated in Fig. \ref{figure_4}, implements our proposed dual-layer feature reconstruction strategy across different network hierarchies. This module reconstructs query features using attention mechanisms applied independently to both last-layer and penultimate-layer feature maps, capturing complementary visual information at multiple abstraction levels.

For $C$-way $K$-shot classification tasks, feature representations are formally defined as follows: $S^1_{(c,k)} \in \mathbb{R}^{r \times d}$ denotes the last-layer feature of the $k$-th support sample in class $c$, with ${S}_{c}^1 \in \mathbb{R}^{kr \times d}$ representing the concatenation of all $\{S^1_{(c,k)}\}_{k=1}^K$ features within that class. Similarly, ${S}_{c}^2 \in \mathbb{R}^{kr \times d}$ denotes the concatenated penultimate-layer features, where $k \in \{1,\ldots,K\}$ and $c \in \{1,\ldots,C\}$. The corresponding query features are represented by $Q_i^1 \in \mathbb{R}^{r \times d}$ and $Q_i^2 \in \mathbb{R}^{r \times d}$, where $i \in \{1,\ldots,C \times M\}$ with $M$ being the number of query samples per class.

These features are then projected into query-key-value spaces through separate linear transformations. For the features from the last layer, ${S}_{c}^1$ and $Q_i^1$ are transformed using the weight matrices $W^1_Q$, $W^1_K$, and $W^1_V \in \mathbb{R}^{d \times d}$ to obtain the projected representations. Specifically, we derive query projections by applying the weight matrices to query features, yielding $Q^1_{(i,Q)}$ from $W^1_Q$ and $Q^1_{(i,V)}$ from $W^1_V$, where the latter is crucial for subsequent distance measurement with reconstructed features. For support features, we apply the respective weight matrices to obtain the key representation $S^1_{(c,K)}$ and value representation $S^1_{(c,V)}$.

For the penultimate layer features, which provide complementary structural details, we apply a separate set of weight matrices $W^2_Q$, $W^2_K$, and $W^2_V \in \mathbb{R}^{d \times d}$ to transform $Q^2_i$ and ${S}_{c}^2$ into their respective query, key, and value projections $Q^2_{(i,Q)}$, $Q^2_{(i,V)}$, $S^2_{(c,K)}$, and $S^2_{(c,V)}$. This dual-pathway projection ensures that both feature hierarchies maintain their distinctive information characteristics throughout the reconstruction process.

We perform attentional reconstruction at both feature levels independently. For the last layer features, we compute the reconstructed query features as follows:
\begin{equation}
	\begin{split}
		\hat{Q}^1_{(c,i)} &= \text{Attention}(Q^1_{(i,Q)}, S^1_{(c,K)}, S^1_{(c,V)}) \\
		&= \sigma\left(\frac{Q^1_{(i,Q)} \left(\sum_{k=1}^{K} S^1_{(c,k)}W^1_K\right)^T}{\sqrt{d}}\right) \sum_{k=1}^{K} \left(S^1_{(c,k)}W^1_V\right),
	\end{split}
\end{equation}
where $\sigma(\cdot)$ represents the softmax function, and $\hat{Q}^1_{(c,i)} \in \mathbb{R}^{r \times d}$ represents the last-layer reconstruction result of the $i$-th query sample from the support set of the $c$-th class.

Similarly, for the penultimate layer features, the reconstructed query features are computed as follows:
\begin{equation}
	\begin{split}
		\hat{Q}^2_{(c,i)} &= \text{Attention}(Q^2_{(i,Q)}, S^2_{(c,K)}, S^2_{(c,V)}) \\
		&= \sigma\left(\frac{Q^2_{(i,Q)} \left(\sum_{k=1}^{K} S^2_{(c,k)}W^2_K\right)^T}{\sqrt{d}}\right) \sum_{k=1}^{K} \left(S^2_{(c,k)}W^2_V\right),
	\end{split}
\end{equation}
where $\hat{Q}^2_{(c,i)} \in \mathbb{R}^{r \times d}$ represents the penultimate-layer reconstruction result of the $i$-th query sample from the support set of the $c$-th class.

The cross-level feature reconstruction architecture processes complementary information from different network hierarchies, enhancing the feature representation capabilities of the model through simultaneous utilization of distinct feature types, which is critical for distinguishing subtle visual differences with limited training examples.

\subsection{Learning Objectives}

\label{Learning Objectives}

After processing through the CLARM module, we obtain the hierarchical reconstructed features for the query sets. The Euclidean distance is then employed to measure the differences between the original query features and their corresponding reconstructed versions for both the last and penultimate layers, as shown in Fig. \ref{figure_2}. The detailed operations are as follows:
\begin{equation}
	\begin{aligned}
		d_{S_c \to Q_i}^1 &= \left\| Q_{(i, V)}^1 - \hat{Q}_{(c, i)}^1 \right\|^2,
	\end{aligned}
	\label{equation_13}
\end{equation}

\begin{equation}
	\begin{aligned}
		d_{S_c \to Q_i}^2 &= \left\| Q_{(i, V)}^2 - \hat{Q}_{(c, i)}^2 \right\|^2.
	\end{aligned}
	\label{equation_14}
\end{equation}
Then, the overall similarity is obtained by calculating the weighted sum of the negative distances:
\begin{equation}
	\begin{aligned}
		s_i^c &= -\tau \left( w_1 \cdot d_{S_c \to Q_i}^1 + w_2 \cdot d_{S_c \to Q_i}^2 \right),
	\end{aligned}
	\label{equation_15}
\end{equation}
where the learnable parameters $w_1$ and $w_2$ adjust the weights of the two feature hierarchy levels, both initialized to $0.5$. This weighting strategy allows the model to adaptively balance information from different network layers. $\tau$ is a learnable temperature factor initialized to $1.0$. The similarity scores are then normalized using the softmax function:
\begin{equation}
	\hat{s}_i^c = \frac{e^{s_i^c}}{\sum_{c=1}^{N} e^{s_i^c}}.
	\label{equation_16}
\end{equation}
For each $N$-way $K$-shot task, the cross-entropy loss based on $\hat{s}_i^c$ is computed as follows:
\begin{equation}
	\mathcal{L} = -\frac{1}{N \times M} \sum_{i=1}^{N \times M} \sum_{c=1}^N \delta(y_i == c) \log(\hat{s}_i^c),
	\label{equation_17}
\end{equation}
where $\delta(y_i == c)$ is 1 if the ground truth label of the query sample matches the $c$-th class, and 0 otherwise. $M$ represents the number of query samples per class.

In the training phase, we refine the proposed network by minimizing $\mathcal{L}$ and consistently apply this process to all randomly generated $N$-way $K$-shot tasks on training dataset $D_{train}$.

\section{Experiments}
\label{Experiments} 
In this section, extensive experiments are conducted to comprehensively evaluate the performance and effectiveness of our proposed method. The experiments are organized as follows:
\begin{itemize}  
	\item Compare the proposed method against state-of-the-art methods on three standard fine-grained datasets using different backbone architectures to establish benchmark performance (Section \ref{Comparison with state-of-the-arts}).
	\item Examine the contribution of individual components through systematic ablation studies, specifically analyzing how the MTFEM and CLARM modules enhance classification performance across various experimental scenarios (Section \ref{Ablation study}).
	\item Investigate the optimal architectural configuration of the MTFEM module by systematically varying Transformer layers and attention heads, analyzing their impact on classification accuracy across different datasets (Section \ref{Impact of Transformer Layer and Head Configurations in MTFEM}).
	\item Provide qualitative evidence through visualization experiments that illustrate the effectiveness of the dual-level feature reconstruction mechanism, demonstrating superior reconstruction capability (Section \ref{Visualization analysis}).
\end{itemize}

\subsection{Datasets}
\label{Datasets}
To evaluate the classification performance of our proposed HMDRN, we conducted experiments on three benchmark datasets widely recognized in fine-grained image classification research. For each dataset, we divided it into $D_{train}$, $D_{val}$ and $D_{test}$. These datasets represent diverse domains of fine-grained classification challenges:

\textbf{CUB-200-2011 (CUB)} \citep{WelinderEtal2010} comprises 11,788 images across 200 distinct bird species. Following established research practices \citep{Zhang_2020_CVPR,ye2020few,chen2019closer}, we utilize human-annotated bounding boxes to crop each image. This dataset follows the standard partition described in \citep{chen2019closer}, with 130 classes for training, 20 classes for validation, and 50 classes for testing.

\textbf{Stanford-Dogs (Dogs)} \citep{xiao2011sun} encompasses 20,580 images of 120 dog breeds. This dataset introduces particular challenges due to significant intra-class variations and the frequent presence of humans alongside dogs in the images. Following the method in \citep{wertheimer2021few}, the data is divided into 70 classes for training, 20 classes for validation, and 30 classes for testing.

\textbf{Stanford-Cars (Cars)} \citep{krause20133d} contains 16,185 images distributed across 196 car models. We use the data split described in \citep{wertheimer2021few}, with 130 classes for training, 17 classes for validation, and 49 classes for testing.

\subsection{Implementation details}
\label{Implementation details}
We employ two widely-adopted backbone networks as feature extractors: Conv-4 \citep{wertheimer2021few} and ResNet-12 \citep{yang2018fewshot}. The Conv-4 architecture comprises four sequential convolutional blocks, with each block integrating 64 filters of size $3 \times 3$, followed by BatchNorm, ReLU activation, and $2 \times 2$ max pooling. When processing $3 \times 84 \times 84$ input images, Conv-4 generates last-layer features of dimension $64 \times 5 \times 5$, while the penultimate layer produces $64 \times 10 \times 10$ feature maps.

For ResNet-12, the architecture consists of four residual blocks, each containing three convolutional layers enhanced with DropBlock regularization. Each convolutional layer incorporates BatchNorm and leaky ReLU activation (negative slope of 0.1), with $2 \times 2$ max pooling applied at the last layer of each block. This backbone transforms $3 \times 84 \times 84$ input images into last-layer features of dimension $640 \times 5 \times 5$ and penultimate-layer features of dimension $320 \times 10 \times 10$.

In the training stage across CUB \citep{WelinderEtal2010}, Dogs \citep{xiao2011sun}, and Cars \citep{krause20133d} datasets, optimization is performed using SGD with Nesterov momentum (0.9) and weight decay ($5 \times 10^{-4}$). A cosine annealing learning rate schedule \citep{loshchilov2017sgdr} is implemented with an initial value of 0.1, and training is conducted for 1200 epochs divided into three equal stages of 400 epochs each. Episodic training is configured with 30-way-5-shot episodes for Conv-4 and 10-way-5-shot episodes for ResNet-12. Model validation is performed every 20 epochs, with the highest-accuracy model being preserved.

The preprocessing pipeline incorporates several data augmentation techniques: random resized cropping, center cropping, horizontal flipping, and color jittering. During feature extraction, complementary information is captured from both the last layer ($5 \times 5$ spatial resolution) and penultimate layer ($10 \times 10$ spatial resolution) of the backbone networks.

The meta-learning system facilitates episode construction with appropriate samplers for both training and testing phases, while cross-entropy loss is utilized to measure discrepancy between predicted class probabilities and ground truth labels during meta-training.

In the evaluation stage, mean accuracy is reported across 10,000 randomly generated 5-way 1-shot and 5-way 5-shot tasks on the test set, with $95\%$ confidence intervals. The hierarchical method integrates similarity scores from both feature levels through learnable weight parameters that are dynamically optimized throughout the training process.

\subsection{Comparison with state-of-the-arts}
\label{Comparison with state-of-the-arts}
\begin{table*}[!ht]
	\centering
	\caption{The 5-way few-shot classification performance on the CUB, Dogs, and Cars datasets using the Conv-4 backbone. The best-performing methods are shown in \textbf{bold}. }
	\label{table_1}
	\resizebox{1.0\textwidth}{!}{
	\begin{tabular}{ccccccc}
		\toprule[1pt]
		\multirow{2}{*}{\it{Method}}      
		& \multicolumn{2}{c}{\it{CUB}} 
		& \multicolumn{2}{c}{\it{Dogs}}  
		& \multicolumn{2}{c}{\it{Cars}}  \\ 
		
		&\multicolumn{1}{c}{$1$-shot} 
		& \multicolumn{1}{c}{$5$-shot}  
		& \multicolumn{1}{c}{$1$-shot}  
		& \multicolumn{1}{c}{$5$-shot}  
		& \multicolumn{1}{c}{$1$-shot} 
		& \multicolumn{1}{c}{$5$-shot} \\ \midrule
		
		Matching~(NeurIPS~16'\citep{DBLP:conf/nips/VinyalsBLKW16})
		& 60.06$\pm$0.88       
		& 74.57$\pm$0.73          
		& 46.10$\pm$0.86    
		& 59.79$\pm$0.72  
		& 44.73$\pm$0.77
		& 64.74$\pm$0.72 \\ 
		
		ProtoNet~(NeurIPS~17'\citep{snell2017prototypical})
		& 64.82$\pm$0.23       
		& 85.74$\pm$0.14          
		& 46.66$\pm$0.21    
		& 70.77$\pm$0.16  
		& 50.88$\pm$0.23
		& 74.89$\pm$0.18 \\ 
		
		Relation~(CVPR 18'\citep{sung2018learning})              
		& 63.94$\pm$0.92          
		& 77.87$\pm$0.64           
		& 47.35$\pm$0.88          
		& 66.20$\pm$0.74            
		& 46.04$\pm$0.91          
		& 68.52$\pm$0.78 \\ 
		
		DN4~(CVPR 19'\citep{8953758})          
		& 57.45$\pm$0.89
		& 84.41$\pm$0.58
		& 39.08$\pm$0.76
		& 69.81$\pm$0.69
		& 34.12$\pm$0.68
		& {87.47$\pm$0.47} \\

		PARN~(ICCV 19'\citep{Wu_2019_ICCV})               
		& {74.43$\pm$0.95}    
		& 83.11$\pm$0.67
		& 55.86$\pm$0.97
		& 68.06$\pm$0.72
		& 66.01$\pm$0.94
		& 73.74$\pm$0.70 \\

		SAML~(ICCV 19'\citep{Hao2019CollectAS})              
		& {65.35$\pm$0.65}    
		& 78.47$\pm$0.41
		& 45.46$\pm$0.36
		& 59.65$\pm$0.51
		& 61.07$\pm$0.47
		& 88.73$\pm$0.49 \\
		
		DeepEMD~(CVPR 20'\citep{Zhang_2020_CVPR})
		& 64.08$\pm$0.50
		& 80.55$\pm$0.71
		& 46.73$\pm$0.49
		& 65.74$\pm$0.63
		& 61.63$\pm$0.27
		& 72.95$\pm$0.38 \\ 
		
		DSN~(CVPR 20'\citep{Zhang_2020_CVPR})
		& 72.56$\pm$0.92
		& 84.62$\pm$0.60
		& 44.52$\pm$0.82
		& 59.42$\pm$0.71
		& 53.15$\pm$0.86
		& 65.19$\pm$0.75 \\ 
		
		MattML~(IJCAI 20'\citep{li2020multiattention})            
		& 66.29$\pm$0.56
		& 80.34$\pm$0.30
		& 54.84$\pm$0.53
		& 71.34$\pm$0.38
		& 66.11$\pm$0.54
		& 82.80$\pm$0.28 \\
		
		CTX~(NeurrIPS 20'\citep{doersch2020crosstransformers})   
		& 72.61$\pm$0.21
		& 86.23$\pm$0.14
		& 57.86$\pm$0.21
		& 73.59$\pm$0.16
		& 66.35$\pm$0.21
		& 82.25$\pm$0.14 \\ 
		
		DLG~(IJMLC 20'\citep{yu2020fewshot})
		& 64.77$\pm$0.90
		& 83.31$\pm$0.55
		& 47.77$\pm$0.86
		& 67.07$\pm$0.72
		& 62.56$\pm$0.82
		& 88.98$\pm$0.47 \\

		LRPABN~(TMM 21'\citep{huang2020low})            
		& 63.63$\pm$0.77
		& 76.06$\pm$0.58
		& 45.72$\pm$0.75
		& 60.94$\pm$0.66
		& 60.28$\pm$0.76
		& 73.29$\pm$0.58 \\

		BSNet(D\&C)~(TIP 21'\citep{li2020bsnet})              
		& 62.84$\pm$0.95
		& 85.39$\pm$0.56
		& 43.42$\pm$0.86
		& 71.90$\pm$0.68
		& 40.89$\pm$0.77
		& 86.88$\pm$0.50 \\

		VFD~(ICCV 21'\citep{zhang2020variational})   
		& 68.42$\pm$0.92
		& 82.42$\pm$0.61
		& 57.03$\pm$0.86
		& 73.00$\pm$0.66
		& -
		& - \\ 
		
		MixtFSL~(ICCV 21'\citep{li2021mixture})   
		& 56.90$\pm$0.88
		& 74.77$\pm$0.75
		& 45.61$\pm$0.78
		& 62.22$\pm$0.68
		& 44.43$\pm$0.79
		& 66.31$\pm$0.75 \\

		FRN~(CVPR 21'\citep{wertheimer2021few})          
		& 74.90$\pm$0.21
		& 89.39$\pm$0.12
		& 60.41$\pm$0.21
		& 79.26$\pm$0.15
		& 67.48$\pm$0.22
		& 87.97$\pm$0.11 \\

		FRN+TDM~(CVPR 22'\citep{lee2022task})
		& 76.55$\pm$0.21        
		& 90.33$\pm$0.11          
		& 62.68$\pm$0.22       
		& 79.59$\pm$0.15         
		& 71.16$\pm$0.21    
		& 89.55$\pm$0.10  \\

		BiFRN~(AAAI 23'\citep{li2023bidirectional})
		& 79.08$\pm$0.20
		& 92.22$\pm$0.10
		& 64.74$\pm$0.22
		& 81.29$\pm$0.14
		& 75.74$\pm$0.20
		& 91.58$\pm$0.09 \\ 
		
		SRNet~(PR 24'\citep{li2024self})
		& 70.79$\pm$0.22
		& 86.42$\pm$0.14
		& 56.00$\pm$0.21
		& 75.48$\pm$0.15
		& 62.04$\pm$0.22
		& 81.12$\pm$0.15 \\
		
		C2-Net~(AAAI 24'\citep{Cross-Layer})
		& 81.00$\pm$0.20
		& 91.31$\pm$0.11
		& 66.42$\pm$0.50
		& 81.23$\pm$0.34
		& \textbf{81.29$\pm$0.45}
		& 91.08$\pm$0.26 \\ \midrule

		Ours
		& \textbf{82.55$\pm$0.19}
		& \textbf{93.50$\pm$0.09}
		& \textbf{66.76$\pm$0.22}
		& \textbf{82.06$\pm$0.14}
		& 77.88$\pm$0.19
		& \textbf{92.80$\pm$0.08}

		\\ \bottomrule
	\end{tabular}
	}
\end{table*}

\begin{table*}[!htp]
	\centering
	\caption{The 5-way few-shot classification performance on the CUB, Dogs, and Cars datasets using the ResNet-12 backbone. The best-performing methods are shown in \textbf{bold}.}
	\label{table_2}
	\resizebox{1.0\textwidth}{!}{
	\begin{tabular}{ccccccc}
		\toprule[1pt]
		\multirow{2}{*}{\it{Method}}      
		& \multicolumn{2}{c}{\it{CUB}} 
		& \multicolumn{2}{c}{\it{Dogs}}  
		& \multicolumn{2}{c}{\it{Cars}}  \\ 
		
		&\multicolumn{1}{c}{$1$-shot} 
		& \multicolumn{1}{c}{$5$-shot}  
		& \multicolumn{1}{c}{$1$-shot}  
		& \multicolumn{1}{c}{$5$-shot}  
		& \multicolumn{1}{c}{$1$-shot} 
		& \multicolumn{1}{c}{$5$-shot} \\ \midrule
		
		ProtoNet~(NeurIPS 17'\citep{snell2017prototypical})      
		& 81.02$\pm$0.20      
		& 91.93$\pm$0.11      
		& 73.81$\pm$0.21      
		& 87.39$\pm$0.12       
		& 85.46$\pm$0.19         
		& 95.08$\pm$0.08 \\

		CTX~(NeurIPS 20'\citep{doersch2020crosstransformers})
		& 80.39$\pm$0.20
		& 91.01$\pm$0.11
		& 73.22$\pm$0.22
		& 85.90$\pm$0.13
		& 85.03$\pm$0.19
		& 92.63$\pm$0.11 \\ 
		
		FEAT~(CVPR 20'\citep{NEURIPS2020_fa28c6cd})
		& 73.27$\pm$0.22
		& 85.77$\pm$0.14
		& -
		& -
		& -
		& - \\ 
		
		DeepEMD~(CVPR 20'\citep{Zhang_2020_CVPR})  
		& 75.59$\pm$0.30
		& 88.23$\pm$0.18
		& 70.38$\pm$0.30
		& 85.24$\pm$0.18
		& 80.62$\pm$0.26
		& 92.63$\pm$0.13  \\ 
		
		RENet~(ICCV 21'\citep{Kang_2021_ICCV})
		& 82.33$\pm$0.44
		& 90.97$\pm$0.26
		& 75.60$\pm$0.46
		& 88.37$\pm$0.25
		& 85.23$\pm$0.40
		& 94.16$\pm$0.20  \\ 
		
		VFD~(ICCV 21'\citep{zhang2020variational})
		& 79.12$\pm$0.83
		& 91.48$\pm$0.39
		& 76.24$\pm$0.87
		& 88.00$\pm$0.47
		& -
		& -  \\

		DeepBDC~(CVPR 22'\citep{li2021joint}) 
		& 81.98$\pm$0.44
		& 92.24$\pm$0.24
		& 73.57$\pm$0.46
		& 86.61$\pm$0.27
		& 82.28$\pm$0.42
		& 93.51$\pm$0.20  \\

		MCL-Katz~(CVPR 22'\citep{wang2021learning})
		& 85.55
		& 93.15
		& 71.49
		& 85.24
		& 84.42
		& 93.32  \\

		MCL-Katz+PSM~(TIP 22'\citep{liu2022learning}) 
		& 85.89
		& 93.08
		& 71.30
		& 85.08
		& 84.72
		& 93.52  \\

		FRN~(CVPR 21'\citep{wertheimer2021few})
		& 84.30$\pm$0.18   
		& 93.34$\pm$0.10         
		& {76.76$\pm$0.21}      
		& {88.74$\pm$0.12}        
		& 88.01$\pm$0.17     
		& 95.75$\pm$0.07 \\

		FRN+TDM~(CVPR 22'\citep{lee2022task})
		& 84.97$\pm$0.18        
		& 93.83$\pm$0.09         
		& 77.94$\pm$0.21      
		& 89.54$\pm$0.12
		& 88.80$\pm$0.16
		& 97.02$\pm$0.06  \\ 
		
		HelixFormer~(ACM MM 22'\citep{zhang2022learning})
		& 81.66$\pm$0.30        
		& 91.83$\pm$0.17         
		& 65.92$\pm$0.49      
		& 80.65$\pm$0.36
		& 79.40$\pm$0.43
		& 92.26$\pm$0.15  \\

		BSFA~(TCSVT 23'\citep{zha2022boosting})
		& 82.27$\pm$0.46        
		& 90.76$\pm$0.26         
		& 69.58$\pm$0.50      
		& 82.59$\pm$0.33
		& 88.93$\pm$0.38
		& 95.20$\pm$0.20  \\ 
		
		ESPT~(AAAI 23'\citep{rong2023espt})
		& 85.45$\pm$0.18
		& 94.02$\pm$0.09
		& 78.26$\pm$0.21
		& 89.35$\pm$0.12
		& 88.20$\pm$0.17
		& 96.96$\pm$0.06  \\

		LCCRN~(TCSVT 23'\citep{li2023locally})
		& 82.97$\pm$0.19        
		& 93.63$\pm$0.10         
		& -      
		& -
		& 87.04$\pm$0.17
		& 96.19$\pm$0.07  \\ 
		
		BiFRN~(AAAI 23'\citep{li2023bidirectional})                   
		& 85.44$\pm$0.18  
		& 94.73$\pm$0.09   
		& 76.89$\pm$0.21  
		& 88.27$\pm$0.12 
		& 90.44$\pm$0.15 
		& 97.49$\pm$0.05 \\ 
		
		SRNet~(PR 24'\citep{li2024self})
		& 83.74$\pm$0.18
		& 93.83$\pm$0.09
		& 75.63$\pm$0.20
		& 88.92$\pm$0.11
		& 84.04$\pm$0.18
		& 93.98$\pm$0.09 \\
		
		C2-Net~(AAAI 24'\citep{Cross-Layer})
		& 78.98$\pm$0.21
		& 89.80$\pm$0.12
		& 75.50$\pm$0.49
		& 87.65$\pm$0.28
		& 88.96$\pm$0.37
		& 95.16$\pm$0.20 \\ \midrule
		
		Ours                    
		& \textbf{86.92$\pm$0.17}   
		& \textbf{95.41$\pm$0.08}   
		& \textbf{78.88$\pm$0.20}  
		& \textbf{89.88$\pm$0.11}  
		& \textbf{90.92$\pm$0.14}
		& \textbf{97.56$\pm$0.05} \\ 
		
		\bottomrule
		
	\end{tabular}
	}
\end{table*}

To validate the effectiveness of our method on fine-grained few-shot image classification tasks, we conduct experiments on the three fine-grained datasets discussed earlier and compare our method with other state-of-the-art methods. The classification accuracy results are shown in Table \ref{table_1} and Table \ref{table_2}.

Conv-4 and ResNet-12 are employed as backbone networks for all comparison methods, and the 5-way 1-shot and 5-way 5-shot classification performance is evaluated. With the Conv-4 backbone, our proposed method demonstrates excellent performance in most scenarios. Specifically, our method shows significant improvements over previous state-of-the-art methods in all cases, except for the 5-way 1-shot setting on the Cars dataset where it is surpassed by C2-Net \citep{Cross-Layer}. With the ResNet-12 backbone, our method exhibits superior performance across all three datasets and scenarios, achieving the highest accuracy among all state-of-the-art methods.

In conclusion, compared to other recently proposed methods, our method achieves consistently outstanding performance on 5-way 1-shot and 5-way 5-shot classification tasks across the three widely used fine-grained image datasets. This success is closely related to the architectural design: the synergistic combination of the MTFEM and CLARM modules, which both leverages complementary features from different network hierarchies to capture comprehensive visual information and precisely focuses on discriminative regions where key differences between visually similar subclasses reside.

\subsection{Ablation study}
\label{Ablation study}
\begin{table*}[!htp]
	\centering
	\caption{Ablation study of the MTFEM and CLARM modules for the 5-way few-shot scheme on the CUB, Dogs, and Cars datasets, with the best performance indicated in \textbf{bold}.}
	\label{table_3}
	\resizebox{1.0\textwidth}{!}{
	\begin{tabular}{ccccccccc}
		\toprule[1pt]
		\multirow{2}{*}{\it{Backbone}} 
		& 
		\multirow{2}{*}{\it{Method}} 
		& \multicolumn{2}{c}{\it{CUB}}
		& \multicolumn{2}{c}{\it{Dogs}} 
		& \multicolumn{2}{c}{\it{Cars}} \\ 
		
		&                             
		& \multicolumn{1}{c}{$1$-shot}
		& \multicolumn{1}{c}{$5$-shot}
		& \multicolumn{1}{c}{$1$-shot}
		& \multicolumn{1}{c}{$5$-shot} 
		& \multicolumn{1}{c}{$1$-shot}
		& \multicolumn{1}{c}{$5$-shot} \\ \midrule
		
		\multirow{7}{*}{Conv-4}
		& 
		Baseline (ProtoNet)       
		& 64.82$\pm$0.23       
		& 85.74$\pm$0.14          
		& 46.66$\pm$0.21       
		& 70.77$\pm$0.16 
		& 50.88$\pm$0.23
		& 74.89$\pm$0.18 \\ 
		
		& 
		(R$^1$)             
		& 76.35$\pm$0.21
		& 90.25$\pm$0.11
		& 61.63$\pm$0.21
		& 80.14$\pm$0.14
		& 71.27$\pm$0.21
		& 88.66$\pm$0.11 \\ 
		
		& 
		(R$^2$)              
		& 71.69$\pm$0.22
		& 87.38$\pm$0.13
		& 52.63$\pm$0.21
		& 70.10$\pm$0.17
		& 61.47$\pm$0.21       
		& 82.32$\pm$0.14 \\ 
		
		& 
		(MTFEM+R$^1$)              
		& 81.67$\pm$0.20
		& 92.49$\pm$0.10
		& 66.75$\pm$0.23
		& 81.71$\pm$0.14
		& 77.47$\pm$0.20      
		& 91.69$\pm$0.09 \\ 
		
		& 
		(MTFEM+R$^2$)              
		& \textbf{83.00$\pm$0.19}
		& 93.30$\pm$0.10
		& 66.44$\pm$0.22
		& 81.79$\pm$0.14
		& 74.98$\pm$0.20    
		& 91.43$\pm$0.09 \\ 
		
		& 
		CLARM(R$^1$+R$^2$)              
		& 75.58$\pm$0.21
		& 90.11$\pm$0.11
		& 61.42$\pm$0.21
		& 80.34$\pm$0.14
		& 71.05$\pm$0.21    
		& 88.82$\pm$0.11 \\ 
		
		&
		(MTFEM+CLARM)
		& 82.55$\pm$0.19
		& \textbf{93.50$\pm$0.09}
		& \textbf{66.76$\pm$0.22}
		& \textbf{82.06$\pm$0.14}
		& \textbf{77.88$\pm$0.19}
		& \textbf{92.80$\pm$0.08} \\ \midrule

		\multirow{7}{*}{ResNet-12}
		& 
		Baseline (ProtoNet)         
		& 81.02$\pm$0.20
		& 91.93$\pm$0.11
		& 73.81$\pm$0.21
		& 87.39$\pm$0.12 
		& 85.46$\pm$0.19         
		& 95.08$\pm$0.08 \\ 
		
		& 
		(R$^1$)
		& 84.26$\pm$0.18
		& 93.78$\pm$0.09
		& 75.86$\pm$0.21
		& 88.36$\pm$0.12
		& 88.10$\pm$0.16 
		& 96.33$\pm$0.07 \\
		
		& 
		(R$^2$)
		& 83.68$\pm$0.18
		& 93.89$\pm$0.09
		& 73.72$\pm$0.21
		& 87.69$\pm$0.12
		& 86.30$\pm$0.16    
		& 95.82$\pm$0.07 \\
		
		& 
		(MTFEM+R$^1$)
		& 86.48$\pm$0.18
		& 94.52$\pm$0.09
		& 77.91$\pm$0.21
		& 89.14$\pm$0.12
		& 88.22$\pm$0.16   
		& 96.04$\pm$0.07 \\
		
		& 
		(MTFEM+R$^2$)
		& 86.02$\pm$0.18
		& 94.59$\pm$0.08
		& 76.78$\pm$0.21
		& 88.46$\pm$0.12
		& 88.58$\pm$0.15   
		& 96.82$\pm$0.06 \\
		
		& 
		CLARM(R$^1$+R$^2$)  
		& 84.74$\pm$0.18
		& 94.33$\pm$0.08
		& 77.13$\pm$0.21
		& 89.31$\pm$0.11
		& 89.90$\pm$0.14      
		& 97.27$\pm$0.06 \\
		
		& 
		(MTFEM+CLARM)
		& \textbf{86.92$\pm$0.17}
		& \textbf{95.41$\pm$0.08}  
		& \textbf{78.88$\pm$0.20}
		& \textbf{89.88$\pm$0.11}
		& \textbf{90.92$\pm$0.14}
		& \textbf{97.56$\pm$0.05}  \\ \bottomrule
	\end{tabular}
	}
\end{table*}

To systematically evaluate the contribution of the MTFEM and CLARM to few-shot fine-grained classification performance, we conduct a series of ablation experiments. Table \ref{table_3} presents the results of these experiments. Specifically, we employ Conv-4 and ResNet-12 as backbone networks to evaluate the classification performance of seven module combinations on 5-way-1-shot and 5-way-5-shot tasks across three fine-grained datasets.
The seven module combinations are as follows:
\begin{itemize}
	\item Baseline (ProtoNet): without MTFEM module and without feature reconstruction on any layer;
	\item (R$^1$): without MTFEM module, with reconstruction only on the last layer features;
	\item (R$^2$): without MTFEM module, with reconstruction only on the penultimate layer features;
	\item (MTFEM+R$^1$): with MTFEM module, with reconstruction only on the last layer features;
	\item (MTFEM+R$^2$): with MTFEM module, with reconstruction only on the penultimate layer features;
	\item CLARM(R$^1$+R$^2$): without MTFEM module, but with reconstruction and fusion of both feature layers;
	\item (MTFEM+CLARM): the complete model with all modules.
\end{itemize}

The experimental results demonstrate that incorporating either of our proposed modules significantly enhances the performance of the baseline ProtoNet. First, comparing R$^1$ and R$^2$ with the baseline, we observe substantial performance gains, indicating that the support set-based query reconstruction method is effective at both network levels. Among these, using only the last layer features (R$^1$) typically achieves higher accuracy than using only the penultimate layer features (R$^2$), which is attributed to the higher semantic presentation present in the last layer features.

When MTFEM is combined with single-layer reconstruction (MTFEM+R$^1$ and MTFEM+R$^2$), we observe significant performance improvements compared to their counterparts without MTFEM (R$^1$ and R$^2$, respectively). This improvement stems from our attention score-based mask-enhanced self-reconstruction strategy, which effectively enhances the ability of the model to perceive subtle differences. The masking mechanism demonstrates consistent improvements across all datasets, confirming its effectiveness for fine-grained tasks that require precise focus on discriminative regions.

Further analysis reveals that CLARM(R$^1$+R$^2$) (which does not use MTFEM but fuses information from both feature levels) outperforms single-layer reconstruction methods, with the advantage over last layer reconstruction being particularly evident with the ResNet-12 backbone. This specific layer selection is grounded in the complementary characteristics of the penultimate and last layers (Fig. \ref{figure_1}). During our model development, attempts to incorporate features from shallower or deeper layers did not lead to improved performance, as they often introduced less discriminative or redundant information under few-shot constraints. Consequently, the benefit of CLARM arises directly from the effective fusion of these two complementary feature levels, which provides a comprehensive visual representation essential for fine-grained classification. The learnable fusion weights further balance these complementary contributions.

Finally, our complete model (MTFEM+CLARM), which integrates both modules, achieves the highest accuracy, outperforming all other combinations. This demonstrates the synergistic effect of combining these two innovations: MTFEM enhances feature discriminability via adaptive binary masking prior to reconstruction, enabling CLARM to operate on noise-reduced representations, while the dual-layer reconstruction of CLARM leverages complementary hierarchical information to maximize the utility of MTFEM-refined features. Both innovations are indispensable and mutually complementary, collectively contributing to performance improvement.

\subsection{Comparison of Binary Mask and Soft Mask Strategies}
\label{Comparison of Binary Mask and Soft Mask Strategies}
\begin{table*}[!htp]
	\centering
	\caption{Performance comparison of binary mask and soft mask strategies for the 5-way few-shot scheme on the CUB, Dogs, and Cars datasets using Conv-4 and ResNet-12 backbones, with the best performance indicated in \textbf{bold}.}
	\label{table_4}
	\resizebox{1.0\textwidth}{!}{
		\begin{tabular}{ccccccccc}
			\toprule[1pt]
			\multirow{2}{*}{\it{Backbone}} 
			& 
			\multirow{2}{*}{\it{Mask Type}} 
			& \multicolumn{2}{c}{\it{CUB}}
			& \multicolumn{2}{c}{\it{Dogs}} 
			& \multicolumn{2}{c}{\it{Cars}} \\ 
			&                             
			& \multicolumn{1}{c}{$1$-shot}
			& \multicolumn{1}{c}{$5$-shot}
			& \multicolumn{1}{c}{$1$-shot}
			& \multicolumn{1}{c}{$5$-shot} 
			& \multicolumn{1}{c}{$1$-shot}
			& \multicolumn{1}{c}{$5$-shot} \\ 
			\midrule
			\multirow{2}{*}{Conv-4}
			& Soft Mask (Sigmoid)       
			& 82.22$\pm$0.19       
			& 93.22$\pm$0.10          
			& 66.44$\pm$0.22       
			& 81.97$\pm$0.14 
			& 77.08$\pm$0.20
			& 92.02$\pm$0.09 \\ 
			& Binary Mask (Ours)              
			& \textbf{82.55$\pm$0.19}
			& \textbf{93.50$\pm$0.09}
			& \textbf{66.76$\pm$0.22}
			& \textbf{82.06$\pm$0.14}
			& \textbf{77.88$\pm$0.19}
			& \textbf{92.80$\pm$0.08} \\ 
			\midrule
			\multirow{2}{*}{ResNet-12}
			& Soft Mask (Sigmoid)         
			& 86.57$\pm$0.17
			& 95.07$\pm$0.08
			& 78.48$\pm$0.20
			& 89.60$\pm$0.11 
			& 89.19$\pm$0.14         
			& 97.38$\pm$0.05 \\ 
			& Binary Mask (Ours)
			& \textbf{86.92$\pm$0.17}
			& \textbf{95.41$\pm$0.08}
			& \textbf{78.88$\pm$0.20}
			& \textbf{89.88$\pm$0.11}
			& \textbf{90.92$\pm$0.14} 
			& \textbf{97.56$\pm$0.05} \\ 
			\bottomrule
		\end{tabular}
	}
\end{table*}

To evaluate the efficacy of the adaptive binary masking mechanism within the Masked Transformer Feature Enhancement Module (MTFEM), we conduct a comparative analysis against a standard continuous soft masking strategy implemented with a sigmoid function. The soft mask produces smooth, differentiable attention weights in the range of zero to one. In contrast, the proposed binary mask applies a hard-thresholding operation that discretizes the attention map by completely suppressing features below an adaptive threshold while preserving those above it. This comparison is designed to elucidate the relative merits of discrete versus continuous attention mechanisms for the few-shot fine-grained image classification task, where the accurate isolation of discriminative regions from background clutter is paramount.

The experimental results, summarized in Table \ref{table_4}, demonstrate a consistent trend. Across both backbone architectures (Conv-4 and ResNet-12) and all three fine-grained datasets, the proposed binary mask outperforms the soft mask in every 5-way 1-shot and 5-way 5-shot setting. This uniform improvement provides strong empirical support for the superiority of the binary masking strategy.

The observed advantage stems from the inherent alignment of the binary mask design with the core requirements of FS-FGIC. As established in the introduction, fine-grained discrimination necessitates the amplification of subtle inter-class differences concentrated in specific regions, while suppressing interference from extensive background noise and intra-class variations. The soft mask, with its continuous weighting scheme, inherently allocates non-zero attention to low-activation background areas. Although this ensures gradient flow, it dilutes the feature representation by blending discriminative signals with non-discriminative components, thereby blurring the decision boundaries between highly similar subclasses—a critical drawback in data-scarce conditions.

Our adaptive binary mask addresses this limitation through its hard-gating mechanism. By applying a threshold derived from the statistics of the attention scores, it partitions the feature space into two discrete states: retained features and suppressed features. This process confers two principal advantages for few-shot fine-grained learning. First, it enhances feature discriminability by generating high-contrast representations that focus exclusively on the most salient regions, thereby sharpening the sensitivity of the model to pivotal inter-class distinctions. Second, it reduces the complexity of the feature space by actively filtering out redundant and noisy elements. This reduction mitigates the risk of overfitting, a crucial consideration in low-data regimes. The resulting sparse and refined feature maps provide a more robust and interpretable foundation for the subsequent hierarchical reconstruction performed by the Cross-Level Attentional Reconstruction Module (CLARM).

In summary, this comparative analysis confirms that the adaptive binary mask constitutes a more effective strategy than the continuous soft mask for FS-FGIC. Its capacity to generate discretized, high-fidelity attention maps is fundamental for the precise localization and enhancement of subtle discriminative features, thereby fulfilling a central objective of the MTFEM within our overall hierarchical dual-reconstruction architecture.

\subsection{Impact of Transformer Layer and Head Configurations in MTFEM}
\label{Impact of Transformer Layer and Head Configurations in MTFEM}
\begin{table*}[!htp]
	\centering
			\caption{Performance comparison of different Transformer layer and head configurations in the MTFEM module for the 5-way few-shot scheme on the CUB, Dogs, and Cars datasets, with the best performance indicated in \textbf{bold}.}
			\label{table_5}
			\resizebox{0.98\textwidth}{!}{
			\begin{tabular}{@{}ccccccccc@{}}  
				\toprule[1pt]
				\multirow{2}{*}{\it{Backbone}}
				&
				\multirow{2}{*}{\it{Layers}}
				&
				\multirow{2}{*}{\it{Heads}}
				& \multicolumn{2}{c}{\it{CUB}}
				& \multicolumn{2}{c}{\it{Dogs}}
				& \multicolumn{2}{c}{\it{Cars}} \\
				
				& & 
				& \multicolumn{1}{c}{$1$-shot}
				& \multicolumn{1}{c}{$5$-shot}
				& \multicolumn{1}{c}{$1$-shot}
				& \multicolumn{1}{c}{$5$-shot}
				& \multicolumn{1}{c}{$1$-shot}
				& \multicolumn{1}{c}{$5$-shot} \\ \midrule
				
				\multirow{8}{*}{Conv-4}
				& 1 & 1
				& 80.07$\pm$0.20       
				& 92.14$\pm$0.10          
				& 65.40$\pm$0.22       
				& 81.80$\pm$0.14
				& 76.40$\pm$0.20
				& 91.48$\pm$0.09 \\
				
				& 1 & 2
				& 80.65$\pm$0.20
				& 92.46$\pm$0.10
				& 66.43$\pm$0.22
				& \textbf{82.26$\pm$0.14}
				& 75.72$\pm$0.20
				& 91.04$\pm$0.09 \\
				
				& 1 & 4
				& 80.77$\pm$0.19
				& 92.54$\pm$0.10
				& 65.40$\pm$0.22
				& 81.51$\pm$0.14
				& 76.65$\pm$0.20
				& 91.82$\pm$0.09 \\
				
				& 2 & 2
				& 81.18$\pm$0.19
				& 92.82$\pm$0.10
				& 66.63$\pm$0.22
				& 82.22$\pm$0.14
				& 77.67$\pm$0.19
				& 92.61$\pm$0.08 \\
				
				& 3 & 1
				& 81.14$\pm$0.19
				& 92.83$\pm$0.10
				& 66.38$\pm$0.22
				& 81.96$\pm$0.14
				& 77.66$\pm$0.19
				& 92.34$\pm$0.09 \\
				
				& 3 & 2
				& \textbf{82.55$\pm$0.19}
				& \textbf{93.50$\pm$0.09}
				& \textbf{66.76$\pm$0.22}
				& 82.06$\pm$0.14
				& \textbf{77.88$\pm$0.19}
				& \textbf{92.80$\pm$0.08} \\
				
				& 3 & 4
				& 82.39$\pm$0.19
				& 93.35$\pm$0.10
				& 66.19$\pm$0.22
				& 81.76$\pm$0.14
				& 76.65$\pm$0.19
				& 92.72$\pm$0.08 \\
				
				& 4 & 4
				& 81.94$\pm$0.19
				& 93.20$\pm$0.10
				& 65.68$\pm$0.22
				& 81.25$\pm$0.14
				& 77.32$\pm$0.20
				& 92.13$\pm$0.09 \\ \midrule

				\multirow{8}{*}{ResNet-12}
				& 1 & 1
				& 86.32$\pm$0.18
				& 94.86$\pm$0.08
				& 77.75$\pm$0.21
				& 89.14$\pm$0.11
				& 90.73$\pm$0.14
				& 97.47$\pm$0.05 \\
				
				& 1 & 2
				& 86.27$\pm$0.17
				& 95.02$\pm$0.08
				& 78.35$\pm$0.20
				& 89.58$\pm$0.11
				& 90.73$\pm$0.14
				& 97.42$\pm$0.06 \\
				
				& 1 & 4
				& 85.90$\pm$0.17
				& 94.77$\pm$0.08
				& 78.26$\pm$0.21
				& 89.18$\pm$0.12
				& 90.82$\pm$0.14
				& 97.48$\pm$0.05 \\
				
				& 2 & 2
				& 86.56$\pm$0.17
				& 95.10$\pm$0.08
				& 78.57$\pm$0.21
				& 89.60$\pm$0.12
				& 90.37$\pm$0.14
				& 97.37$\pm$0.05 \\
				
				& 3 & 1
				& 86.62$\pm$0.17
				& 95.25$\pm$0.08
				& 76.28$\pm$0.21
				& 88.36$\pm$0.12
				& 90.39$\pm$0.14
				& 97.44$\pm$0.05 \\
				
				& 3 & 2
				& \textbf{86.92$\pm$0.17}
				& \textbf{95.41$\pm$0.08}
				& \textbf{78.88$\pm$0.20}
				& \textbf{89.88$\pm$0.11}
				& \textbf{90.92$\pm$0.14}
				& \textbf{97.56$\pm$0.05} \\
				
				& 3 & 4
				& 86.51$\pm$0.17
				& 95.03$\pm$0.08
				& 77.86$\pm$0.21
				& 89.31$\pm$0.11
				& 89.94$\pm$0.14
				& 97.43$\pm$0.05 \\
				
				& 4 & 4
				& 86.67$\pm$0.17
				& 95.09$\pm$0.08
				& 77.86$\pm$0.20
				& 89.27$\pm$0.11
				& 89.89$\pm$0.14
				& 97.26$\pm$0.05 \\ \bottomrule
			\end{tabular}
		}
\end{table*}
The configuration of the Transformer structure in the MTFEM module directly influences the ability of the model to process fine-grained visual features. The number of layers determines the depth and abstraction level of feature transformation, while the number of heads affects the capacity of the model to capture different types of feature relationships by dividing the feature space into multiple subspaces. Considering the critical role of the mask-enhanced Transformer self-reconstruction mechanism in model performance, we systematically investigate the impact of different layer and head configurations on classification performance.

As shown in Table \ref{table_5}, we design a series of experiments, systematically adjusting the number of Transformer layers (1 to 4) and attention heads (1, 2, 4) in the MTFEM module while keeping other model parameters unchanged. All experiments are conducted across three fine-grained datasets using both Conv-4 and ResNet backbone networks, testing both 5-way 1-shot and 5-way 5-shot settings.

Experimental results demonstrate that the layer and head configurations of the Transformer significantly impact model performance. Single-layer Transformer configurations, regardless of head count, limit the ability of the model to capture complex feature relationships, indicating that the mask enhancement mechanism requires sufficient layers to gradually refine feature representations. Conversely, overly complex configurations lead to performance degradation due to overfitting or over-parameterization of the limited support samples. Our optimal configuration (layers=3, heads=2) achieves an ideal balance, delivering the best performance in almost all experimental settings.

The effectiveness of our optimal configuration can be analyzed through the core mechanisms of the MTFEM module. Three Transformer layers provide sufficient depth for the mask enhancement mechanism to refine features at multiple abstraction levels, initially identifying and subsequently strengthening the feature representations of discriminative regions. Meanwhile, two attention heads enable the capture of different feature relationships while avoiding attention dispersion issues. Our core innovation, the mask enhancement mechanism, is gradually refined through the 3-layer structure, focusing attention more precisely on the most discriminative regions through multi-layer mask processing. Additionally, this configuration provides sufficient model capacity in a few-shot learning environment while avoiding the decrease in generalization capability caused by excessive parameterization.

In summary, the optimal configuration of the MTFEM module reflects the best balance between model complexity and task requirements, fully leveraging the advantages of the mask enhancement mechanism in few-shot fine-grained classification through a moderately deep and appropriately multi-headed structure.

\subsection{Visualization analysis}
\label{Visualization analysis}
\begin{figure*}[ht]
	\centering
	\includegraphics[width=1.0\linewidth]{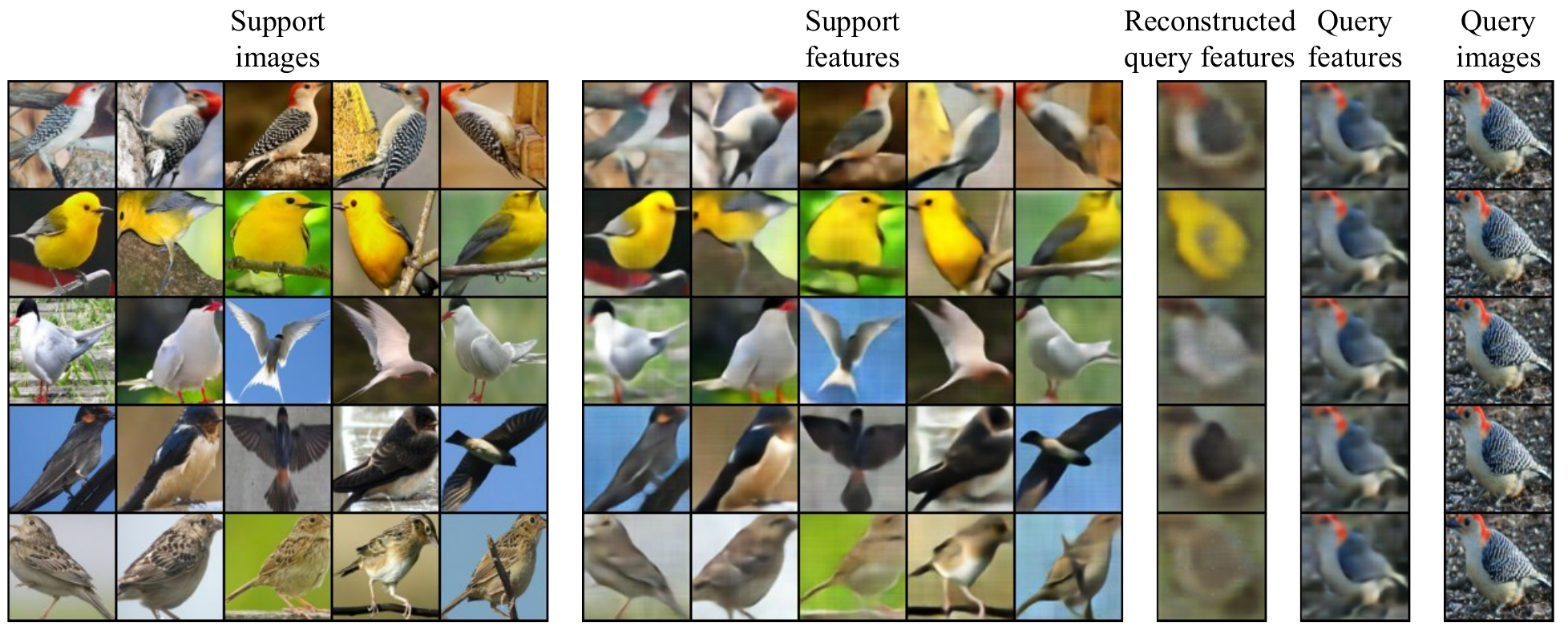}
	\caption{Visualization of original and reconstructed features from the penultimate layer of our model on the CUB dataset.}
	\label{figure_5}
\end{figure*}
\begin{figure*}[ht]
	\centering
	\includegraphics[width=1.0\linewidth]{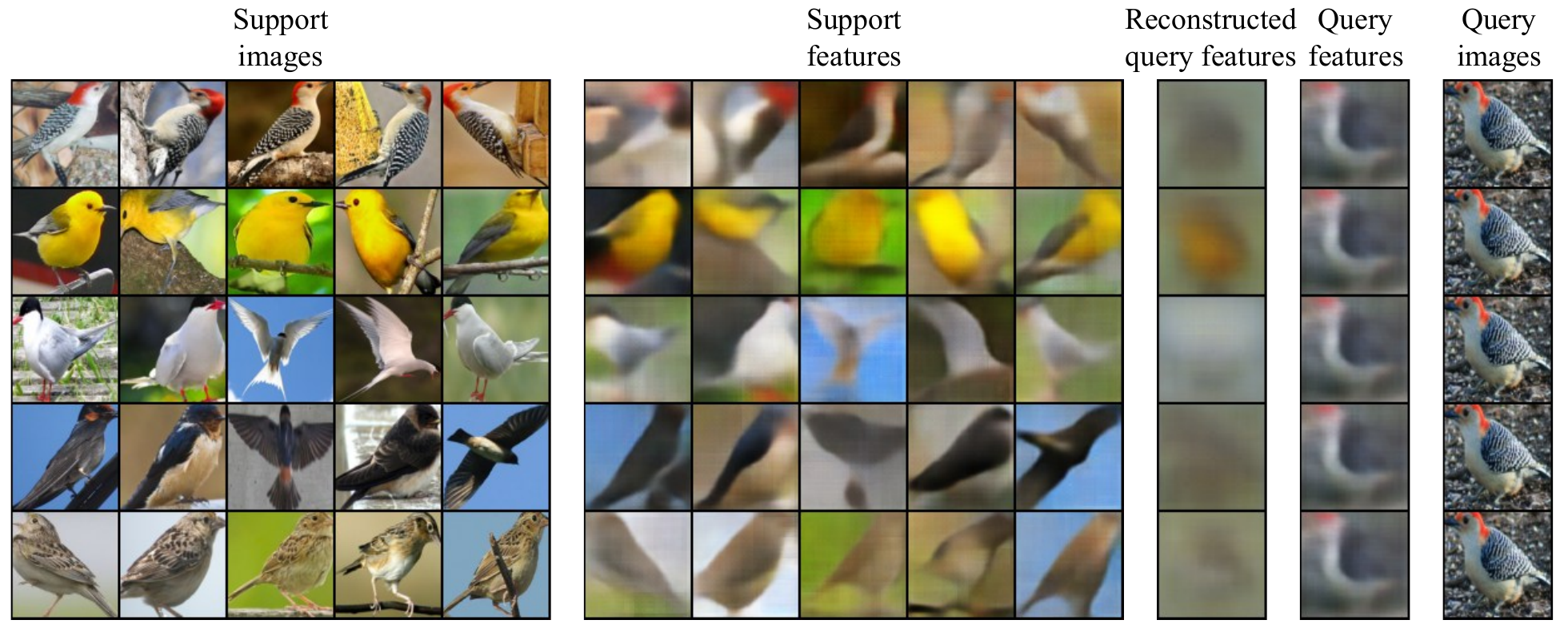}
	\caption{Visualization of original and reconstructed features from the last layer of our model on the CUB dataset.}
	\label{figure_6}
\end{figure*}
To visually demonstrate the superior performance of our proposed method in feature reconstruction, we recover and visualize both original and reconstructed features for comparative analysis. First, we train a reverse ResNet decoder to map features back to original images, enabling intuitive comparison of reconstruction effects across different models. Specifically, we use the Adam optimizer with an L1 loss function, initial learning rate of 0.01, batch size of 200, and train for 500 epochs. For the last layer features, the decoder transforms $640 \times 5 \times 5$ features into $3 \times 84 \times 84$ three-channel images; for the penultimate layer features, it converts $320 \times 10 \times 10$ features into images of the same dimensions.

We obtain outputs from the penultimate and last layers of feature reconstruction and visualize them respectively, as shown in Fig. \ref{figure_5} and Fig. \ref{figure_6}. The leftmost column displays support set images from five different classes, with each class occupying one row, while the query image is repeated five times in the rightmost column. The second column from the left shows the recovered images of support features $S^1_{(c,V)}$ and $S^2_{(c,V)}$. Similarly, the second column from the right shows the recovered images of query features $Q^1_{(i,V)}$ and $Q^2_{(i,V)}$. The third column from the right presents the recovered images of the reconstructed query features $\hat{Q}^1_{(c,i)}$ and $\hat{Q}^2_{(c,i)}$ of our model based on support features from each class.

When the support set and query set belong to the same class, the reconstructed query features $\hat{Q}^2_{(c,i)}$ exhibit extremely high similarity to the original query features $Q^2_{(i,V)}$; whereas when they come from different classes, the similarity significantly decreases. This observation directly validates the effectiveness of our reconstruction mechanism in capturing class-specific discriminative features.

The visualization results demonstrate the superior feature reconstruction capability of our model. Notably, the reconstruction quality of penultimate layer features $\hat{Q}^2_{(c,i)}$ surpasses that of the last layer $\hat{Q}^1_{(c,i)}$, highlighting the dynamic integration of hierarchical features during training. Analysis of learned weights $w_1$ and $w_2$ reveals dataset-dependent integration strategies. For CUB and Cars datasets, the penultimate layer receives greater emphasis ($w_2 > w_1$), leveraging its richer spatial details for fine-grained discrimination. Conversely, for the Dogs dataset, the weight distribution shifts toward the last layer, suggesting greater reliance on high-level semantic features to accommodate pronounced intra-class variation. This adaptive behavior confirms that the penultimate layer generally provides crucial fine-grained spatial information, whereas the last layer offers complementary semantic representation. Our method automatically optimizes the fusion of these complementary representations, demonstrating flexibility across diverse fine-grained classification tasks.

These visualization results validate the excellent performance of our model in capturing discriminative visual features and achieving precise feature reconstruction, visually demonstrating its clear advantage in reconstruction quality.

\section{Discussion}
\label{Discussion}
\begin{table}[!htp]
	\centering
	\caption{Comparison of model complexity and computational cost of FRN, BiFRN, and the proposed method on the CUB, Dogs, and Cars datasets using the ResNet‑12 backbone.}
	\label{table_6}
	\resizebox{0.7\textwidth}{!}{
		\begin{tabular}{cccccc}
			\toprule[1pt]
			
			& \multirow{2}{*}{\it{Method}} 
			& \multirow{2}{*}{\it{Number of parameters (M)}} 
			& \multicolumn{3}{c}{\it{Training time per epoch}}  \\ 
			
			& 
			&                         
			& \multicolumn{1}{c}{CUB}    
			& \multicolumn{1}{c}{Dogs}
			& \multicolumn{1}{c}{Cars}       \\ \midrule
			
			& 
			FRN     
			& 12.42
			& 9.82 s
			& 15.37 s
			& 14.62 s \\ 
			
			& 
			BiFRN     
			& 16.13
			& 11.29 s
			& 17.45 s
			& 15.71 s \\ 
			
			& 
			HMDRN (Ours)        
			& 23.25
			& 13.26 s
			& 20.95 s
			& 18.47 s\\ \bottomrule
		\end{tabular}
	}
\end{table}
\subsection{Computational Cost}
\label{Computational Cost}
We assess the computational cost of the proposed method by comparing the total number of network parameters and the average training time per epoch, as detailed in Table~\ref{table_6}. All experiments were implemented in PyTorch on an NVIDIA RTX 3090 GPU. HMDRN contains 23.25\,M parameters,  representing increases of approximately $87\%$ over FRN (12.42\,M) and $44\%$ over BiFRN (16.13\,M). This rise is attributable primarily to the introduced Cross-Level Attentional Reconstruction Module (CLARM) and the Masked Transformer Feature Enhancement Module (MTFEM), which are essential for capturing hierarchical and discriminative features. In terms of training time, HMDRN requires 13.26\,s, 20.95\,s, and 18.47\,s per epoch on the CUB, Dogs, and Cars datasets, respectively. While these figures are higher than those of the baseline methods, the associated increase in training cost remains moderate and is justified by the significant performance gains demonstrated in our experiments.

\subsection{Limitations of our method}
\label{Limitations of our method}
Despite the promising classification performance of our proposed method, several limitations need to be acknowledged. First, while our model demonstrates competitive performance on established fine-grained datasets, its generalization to cross-domain scenarios \citep{ding2026fa} remains unexamined. Second, the evaluation of our model is currently limited to natural scene images. Its applicability to real-world scenarios such as industrial quality inspection or autonomous driving remains untested. Finally, the method leads to increased computational cost during training, which may restrict its deployment in resource-constrained environments.

\subsection{Future work}
\label{Future work}
To address the identified limitations and expand the utility of the HMDRN framework, several research directions merit further exploration. Future work should examine cross‑domain generalization capabilities, evaluating performance under distribution shifts and developing lightweight adaptation strategies to improve robustness in unseen environments.\citep{ding2025decoupling} Computational efficiency could be enhanced through the adoption of efficient architecture designs to maintain accuracy while reducing resource demands. Extending the framework to multimodal and sequential data inputs would broaden its applicability to domains such as biomedical imaging and environmental monitoring. \citep{ding2025person} Finally, integrating interpretability mechanisms could help clarify the decision process of the model, particularly for applications requiring high transparency or operational safety.

\section{Conclusion}
\label{Conclusion}
In this paper, we propose a hierarchical mask-enhanced dual reconstruction network (HMDRN). It consists of a mask-enhanced self-reconstruction module and a dual-layer feature reconstruction module, both of which effectively leverage complementary visual information from different feature hierarchies and enhance focus on discriminative regions. Comprehensive experiments on three challenging fine-grained datasets demonstrate that HMDRN consistently outperforms existing state-of-the-art methods. These results underscore the ability of our method to successfully address the core challenges in FS-FGIC by precisely identifying discriminative regions while simultaneously maintaining robust generalization capabilities with limited training examples.

\section*{Acknowledgements}
This work was supported by the National Natural Science Foundation of China (Grant No.  62566040, 62106090,62276270), the Natural Science Foundation of Jiangxi Province (Grant No. 20232BAB212027, 20242BAB23013), and the Thousand Talents Plan of Jiangxi Province (Grant No. jxsg2023101085).

\bibliographystyle{elsarticle-harv}
\bibliography{elsarticle}

\end{document}